\documentclass[conference,a4paper]{IEEEtran}
\usepackage{amsmath,amssymb,amsfonts,amsthm}
\usepackage{setspace}
\usepackage{support-caption}
\usepackage{subcaption}
\usepackage{algorithmic}
\usepackage{graphicx}
\usepackage{pgfplots}
\usepackage{siunitx}
\usepackage{textcomp}
\usepackage{xcolor}
\usepackage{setspace}
\usepackage{float}
\usepackage{makecell}
\usepackage{hyperref}
\usepackage[numbers, square, comma, compress]{natbib}
\usepackage{tikz}
\usetikzlibrary{math,shapes}
\def\BibTeX{{\rm B\kern-.05em{\sc i\kern-.025em b}\kern-.08em
    T\kern-.1667em\lower.7ex\hbox{E}\kern-.125emX}}
\pgfplotsset{compat=1.17}

\makeatletter
\let\old@ps@headings\ps@headings
\let\old@ps@IEEEtitlepagestyle\ps@IEEEtitlepagestyle
\def\confheader#1{%
\def\ps@headings{%
\old@ps@headings%
\def\@oddhead{\strut\hfill#1\hfill\strut}%
\def\@evenhead{\strut\hfill#1\hfill\strut}%
}%
\def\ps@IEEEtitlepagestyle{%
\old@ps@IEEEtitlepagestyle%
\def\@oddhead{\strut\hfill#1\hfill\strut}%
\def\@evenhead{\strut\hfill#1\hfill\strut}%
}%
\ps@headings%
}
\makeatother

\confheader{%
2022 International Conference on Indoor Positioning and Indoor Navigation (IPIN), 5 - 7 Sep. 2022, Beijing, China}

\IEEEoverridecommandlockouts
\IEEEpubid{\makebox[\columnwidth]
{\hfill 978-1-7281-6218-8/22/\$31.00~\copyright2022~IEEE}
\hspace{\columnsep}\makebox[\columnwidth]{ }}

\begin{document}

\newtheorem{definition}{Definition}
\newtheorem{example}{Example}
\newcommand{\comment}[1]{}

\title{Decision Trees for Analyzing Influences on the Accuracy of Indoor Localization Systems}

\makeatletter
\newcommand{\linebreakand}{%
  \end{@IEEEauthorhalign}
  \hfill\mbox{}\par
  \mbox{}\hfill\begin{@IEEEauthorhalign}
}
\makeatother

\author{\IEEEauthorblockN{1\textsuperscript{st} Jakob Schyga}
\IEEEauthorblockA{\textit{Institute for Technical Logistics} \\
\textit{Hamburg University of Technology}\\
Hamburg, Germany \\
jakob.schyga@tuhh.de}
\and
\IEEEauthorblockN{2\textsuperscript{nd} Swantje Plambeck}
\IEEEauthorblockA{\textit{Institute of Embedded Systems} \\
\textit{Hamburg University of Technology}\\
Hamburg, Germany \\
swantje.plambeck@tuhh.de}
\and
\IEEEauthorblockN{3\textsuperscript{rd} Johannes Hinckeldeyn}
\IEEEauthorblockA{\textit{Institute for Technical Logistics} \\
\textit{Hamburg University of Technology}\\
Hamburg, Germany \\
johannes.hinckeldeyn@tuhh.de}
\linebreakand
\IEEEauthorblockN{4\textsuperscript{th} Görschwin Fey}
\IEEEauthorblockA{\textit{Institute of Embedded Systems} \\
\textit{Hamburg University of Technology}\\
Hamburg, Germany \\
goerschwin.fey@tuhh.de}
\and
\IEEEauthorblockN{5\textsuperscript{th} Jochen Kreutzfeldt}
\IEEEauthorblockA{\textit{Institute for Technical Logistics} \\
\textit{Hamburg University of Technology}\\
Hamburg, Germany \\
jochen.kreutzfeldt@tuhh.de}
}

\maketitle

\renewcommand{\baselinestretch}{0.97}
\captionsetup{font=small,skip=5pt,belowskip=0pt}
\setlength{\textfloatsep}{4pt}
\setlength{\dbltextfloatsep}{5pt}

\begin{abstract}

Absolute position accuracy is the key performance criterion of an Indoor Localization System (ILS). Since ILS are heterogeneous and complex cyber-physical systems, the localization accuracy depends on various influences from the environment, system configuration, and the application processes. To determine the position accuracy of a system in a reproducible, comparable, and realistic manner, these factors must be taken into account. We propose a strategy for analyzing the influences on the position accuracy of ILS using decision trees in combination with application-related or technology-related categorization. The proposed strategy is validated using empirical data from 120 experiments. The accuracy of an Ultra-Wideband and a LiDAR-based ILS was determined under different application-driven influencing factors, considering the application of autonomous mobile robots in warehouses. Finally, the opportunities and limitations of analyzing decision trees to compare system performance, find a suitable system, optimize the environment or system configuration, and understand the relevance of different influencing factors are presented.
\end{abstract}


\section{Introduction}
\label{sec:intro}

Knowing the position of an entity is essential for a multitude of applications such as equipment tracking in hospitals~\cite{9152857}, elderly people support~\cite{s21103549}, augmented reality~\cite{8377831}, asset tracking in warehouses~\cite{s20133709}, or autonomous mobile robots~\cite{kusuma2019designing}. Indoor Localization Systems (ILS) enable the determination of an entity's position in indoor environments. In today's industry, ILS are of particular importance to meet the increasing demands for efficiency, transparency, flexibility, and safety.

ILS are highly heterogeneous and complex cyber-physical systems. ~\citet{Reinke.092013} present strategies for contour-based self-localization based on LiDAR,~\citet{Kalaitzakis.2021} present an experimental comparison of camera-based pose estimation with fiducial markers, and~\citet{9662610} present an approach for machine learning-based localization using visible light. All these examples show that the performance of ILS depends on a multitude of influences. 

Figure~\ref{fig:intro} presents an application-driven perspective on the performance of ILS under influences. A real-world application or application domain determines the environment and application processes. An ILS is deployed in this environment in a particular soft- and hardware configuration. The system's performance depends on the system configuration, the environment, and the processes that determine the movement of an entity to be localized. To enable an application, the system performance must be higher than required by the application. Performance requirements are derived by analyzing the application processes. 

\begin{figure}[b]
    \centering
    \includegraphics[width=0.4\textwidth]{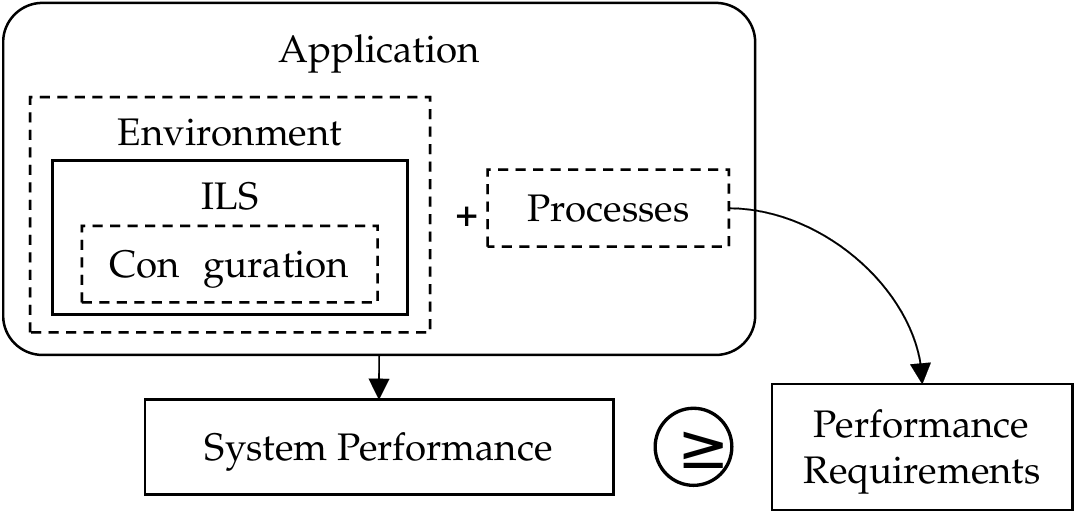}
    \caption{Application-Driven Perspective on ILS Performance}
    \label{fig:intro}
\end{figure}

Test and Evaluation (T\&E) of ILS serves system developers and users to determine and compare system performance, understand system behavior, optimize the system or environment configuration, and find a suitable system for an application. The absolute accuracy of a localization estimate is considered the key performance criterion of a localization system. Since ILS are heterogeneous and complex cyber-physical systems, the position accuracy can depend on various influencing factors. 

ILS are commonly tested and evaluated in benchmarking studies~\cite{9662632, 7217149}, indoor localization competitions~\cite{Potorti.102015, Lymberopoulos.2015}, and in system development~\cite{9662528, 9662520}. However, systematic analysis of influencing factors is rare. To achieve meaningful test and evaluation results, the application-driven perspective as presented in Figure~\ref{fig:intro} must be taken into account.

In this work, we propose a strategy for the analysis of influences on the absolute position accuracy of ILS using decision trees. For the analysis of the T\&E results, decision trees are learned in combination with application-related and technology-related categorization. The benefits of decision trees are rooted in their simple structure and intuitive interpretability.
Related research is presented in Section~\ref{sec:rel}. Section~\ref{sec:methods} describes the strategy for identifying influencing factors, determining the system's accuracy, and learning decision trees. The proposed strategy is applied to an Ultra-Wideband (UWB)-based and a Light Detection and Ranging (LiDAR)-based ILS, as presented in Section~\ref{sec:exp} under consideration of an industrial application. In Section~\ref{sec:discussion}, the usefulness of a decision tree-based analysis is evaluated for several typical problems in the practical use of localization systems. In addition, limitations are discussed. Finally, we provide conclusions in Section~\ref{sec:conclusion}.
\section{Related Work}
\label{sec:rel}

Approaches for T\&E can be divided into system vs. component-testing, black-box vs. white-box (or grey-box) testing, or building-wide vs. laboratory testing~\cite{ISOIECJTC1SC31.2016, s22072797}. Influences can be considered for the experiment design and/or for analysis. In system-level testing, influences are commonly considered by designing the system set-up, the choice of the building, and the entity-type appropriately for the localization system and application under consideration. For example,~\citet{HAMMER201561} test ILS for miners in an underground mine and~\citet{6937007} for Automated Guided Vehicles (AGVs) in warehouses. Occasionally, individual influences are examined. For example,~\citet{s20174922} analyze the influence of image noise on the accuracy of a visual Simultaneous  Localization and Mapping (vSLAM) algorithm, while~\citet{6613917} examine influences on the received signal strength for Wi-Fi-based localization. For comparison, results are commonly visualized by cumulative distribution functions or boxplots~\cite{8738808}.
However, determining the magnitude of various influencing factors on the accuracy of an ILS is rare. 

Methodologies for T\&E of indoor localization systems on a system-level exist and can be applied to systematize the process of test and evaluation. As for the consideration of influences, the EvAAL Framework only requires the building, the path, and the entity to be localized to represent the considered application~\cite{Potorti.2017}, while the ISO/IEC 18305~\cite{ISOIECJTC1SC31.2016} additionally suggests to include particularly challenging experiments for the system under test. The T\&E~4iLoc Framework~\cite{s22072797} proposes a procedure to systematically define application-driven influencing factors and transpose them into an experiment in a semi-controlled test environment such as a test hall. Finally, the EVARILOS Benchmarking Handbook~\cite{vanHaute.2013} proposes the systematic analysis of influences from changes in the environment, mobility, amount of radio-frequency nodes, or radio interference to a reference scenario individually. Hence, the EVARILOS Benchmarking Handbook~\cite{vanHaute.2013} is the only methodology, that directly includes methods to analyze influencing factors. However, the determination of such sensitivity values comes with the following limitations:

\begin{itemize}
    \item influences are only considered as a comparison to a reference scenario,
    \item dependencies between influencing factors are not considered,
    \item non-linear dependencies are neglected.
\end{itemize}

Therefore, sensitivity values are not suitable for describing a system's accuracy under various complex influences. An integrative approach for the application-driven determination and analysis of influences does not yet exist.
\section{Learning Decision Trees on Influences to the Position Accuracy}
\label{sec:methods}

In the following, we propose a strategy for application-driven T\&E of indoor localization systems to determine the position accuracy under different influencing factors. The problem solved by this strategy is to decide which localization performance is achievable under given influencing factors. Due to their simple structure and intuitive interpretability, we use decision trees to represent the T\&E results. A decision tree is a graph-based representation of a decision function. Given a set of influences, the decision tree decides which localization performance is achieved. Figure~\ref{fig:process} depicts the three steps of the proposed strategy:
\begin{enumerate}
    \item[A.] Performance metrics of the system under test are determined for each experiment together with an associated set of influencing factors.
    \item[B.] The resulting performance metrics for each experiment are categorized.
    \item[C.] The influencing factors labeled with performance classes are used to learn a decision tree.
\end{enumerate}

\begin{figure}[b]
    \centering
    \includegraphics[width=0.99\linewidth]{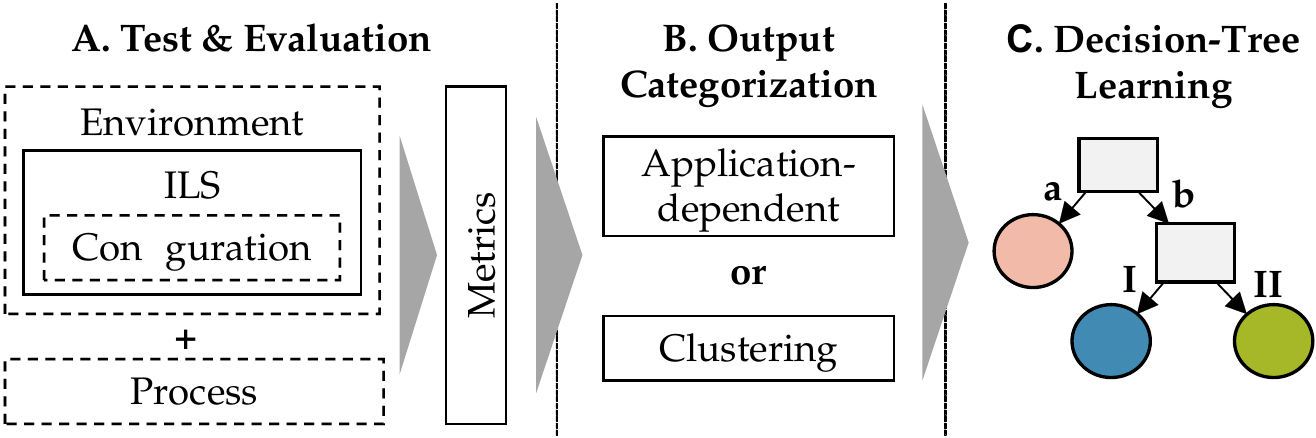}
    \caption{Proposed Strategy for Learning Decision Trees}
    \label{fig:process}
\end{figure}

\subsection{Test and Evaluation}

In step A of Figure~\ref{fig:process}, performance metrics are collected from experiments under several different scenarios.
To ensure repeatability of results for the complex experiment scenario, the T\&E~4iLoc Framework~\cite{s22072797} is utilized as methodology.


With the T\&E 4iLoc Framework~\cite{s22072797}, test and evaluation works as follows. An application or application domain defines a scenario. A scenario specifies influencing factors for an experiment, such as the type of entity to be localized, the lighting conditions, and static or dynamic objects in the environment. The experiment is set up according to the scenario and an entity to be localized is equipped with the localization and a reference system for continuous localization. The coordinate systems are aligned by applying the Umeyama-Alignment~\cite{Umeyama.1991}. The experiment is then executed by sequentially passing a number of predefined evaluation poses. Finally, by comparing the ground truth position with the system's location estimate at the evaluation poses, various accuracy metrics are computed. 
The accuracy metrics are associated with the influencing factors of the scenario. We repeat each experiment several times to investigate the significance of the influencing factors.
The average of the performance metrics over all experiments from the same scenario is finally used for the further analysis steps.

\subsection{Output Categorization}

The representation in decision trees, discussed in the next subsection, requires categorization of the performance metric. The first possibility is an application-related categorization where boundaries are defined manually by setting performance classes based on requirements of an application.

If the specific application or requirements are not known or not relevant, automatic clustering can provide reasonable performance classes. Automatically generated clusters are more related to the capabilities of the system and thus represent a technology-related categorization. The k-nearest neighbor (kNN) clustering method is proposed as a well-established approach. When clustering with kNN, the number of clusters must be determined a priori. The elbow criterion is used to find the best tradeoff between clustering accuracy and overfitting, while the clustering inaccuracy in the case of kNN is measured by the sum of squared distances of the data points and their cluster center~\cite{elbow-criterion}.

\subsection{Decision Tree Learning}

A decision tree allows the systematic analysis of various features for a single output metric. Here, we use the performance metric as the output while the influencing factors represent the input features.
To learn a decision tree, labeled feature vectors are required. Here, this is the set of influences of a T\&E scenario associated with the respective performance class. During the learning process, individual decisions are chosen so that all feature vectors leading to the same leaf have the same categorical label in the learning set. Knowing a feature vector, we can follow the decision steps from the entry node (called \textit{root}) to an output category (encoded in the \textit{leafs}) to obtain the category corresponding to the given feature vector. 

\section{Experiments and Results} 
\label{sec:exp}

In this section, the proposed strategy to learn decision trees for the analysis of the system performance of ILS under influences is applied to an empirical case-study. To obtain a realistic experiment setup, influencing factors, and output categorization, an application must be considered. The application of Autonomous Mobile Robots (AMRs) is chosen, because AMRs require knowledge of their position for various tasks such as navigation, obstacle avoidance, material handling, or goods picking. UWB and LiDAR are common technologies for locating AMRs in warehouses~\cite{9460266, 9494634}. We present the procedure from the determination of influencing factors to learned decision trees in the following for an UWB-based ILS (LOCU~\cite{LOCU}, SICK AG) and a LiDAR-based ILS. The LiDAR ILS consists of a multilayer LiDAR scanner (microScan3~\cite{ms3}), a control unit (SIM1000~\cite{sim1000}), and a localization software (LiDAR-LOC~\cite{LLS}) from SICK AG.

\subsection{Experimental Setup and Factors of Influence}

In the following, the experimental setup, parameters, and technologies are introduced. First, an overview of the test environment is given. Afterwards, details on the two ILS, namely LiDAR and UWB, are given together with their influencing factors that exist in the test environment.

\subsubsection{Test Environment and Experiment Setup}
The experiments are  carried out at a test facility at the TUHH Institute for Technical Logistics. The facility contains various objects such as racks, load carriers, and industrial vehicles and thus represents a typical warehouse environment. A high-precision optical motion capture system provides localization data that serves as a position reference for accuracy evaluation and robot control~\cite{article_mocap}. The collected performance measure, thus, is the euclidian distance between the location estimate of an ILS and the actual position provided by the reference system. The TurtleBot2 robot platform is equipped with the localization devices to be used as the entity to be localized (Figure~\ref{fig:setupa}). 

\begin{figure*}[th]
\centering
\begin{subfigure}{.32\textwidth}
  \centering
  \includegraphics[width=\linewidth]{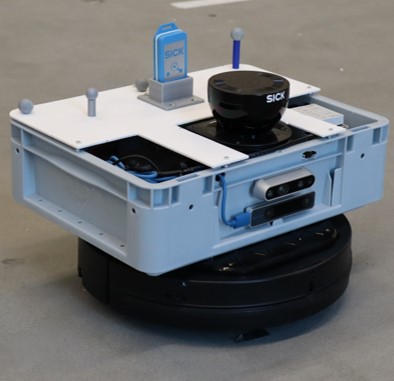}
  \subcaption{}
  \label{fig:setupa}
\end{subfigure}
\begin{subfigure}{0.232\textwidth}
  \centering
  \includegraphics[width=\linewidth]{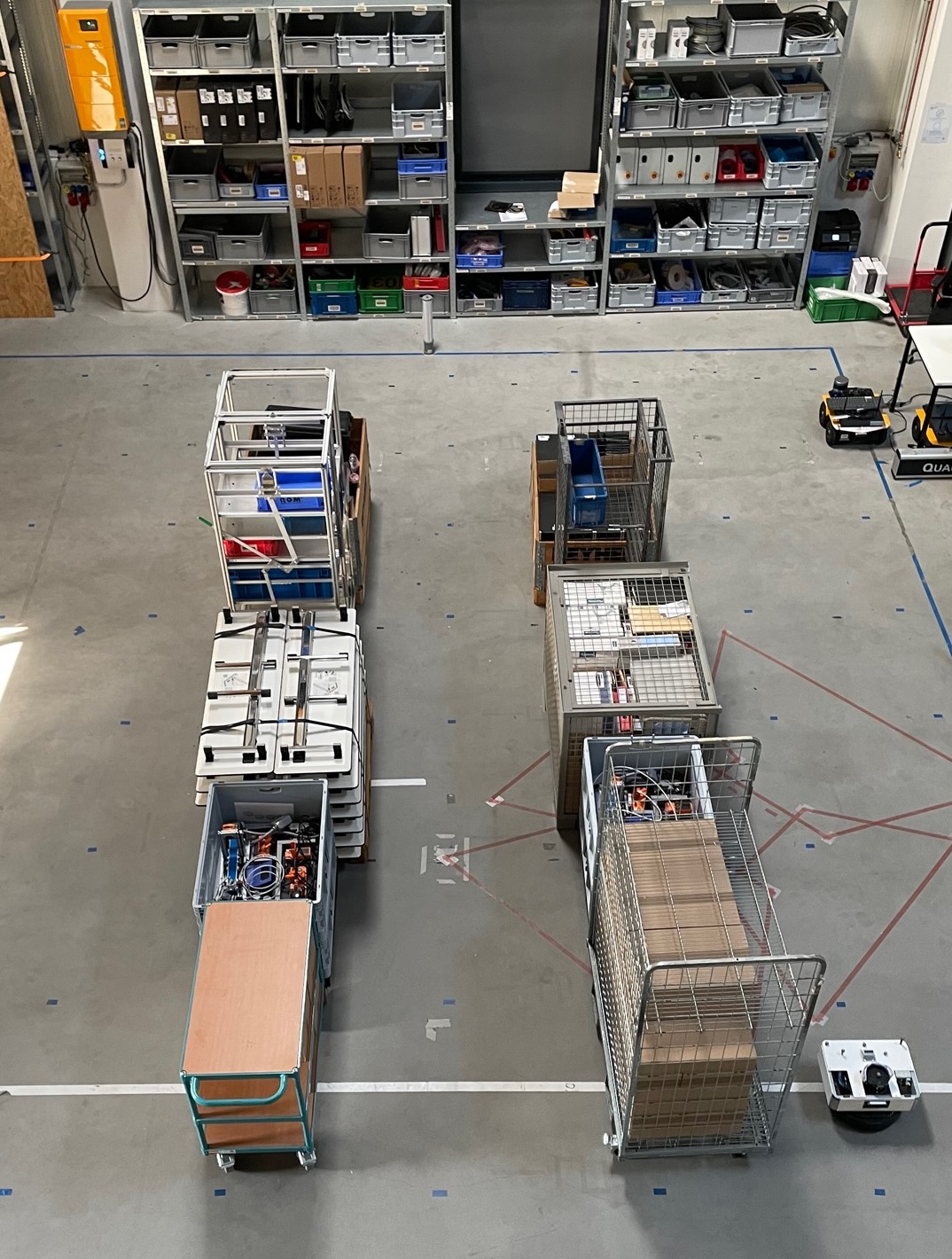}
  \subcaption{}
  \label{fig:setupb}
\end{subfigure}
\begin{subfigure}{.391\textwidth}
  \centering
  \includegraphics[width=\linewidth]{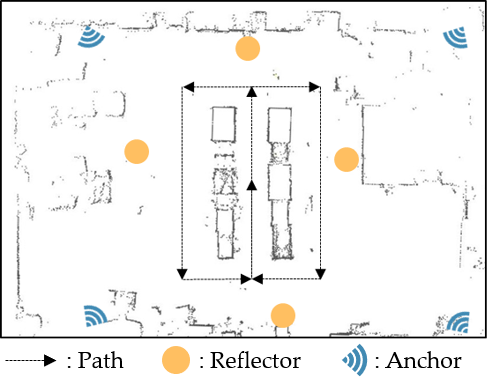}
  \subcaption{}
  \label{fig:setupc}
\end{subfigure}
\caption{Experimental Setup: (a) TurtleBot2 equipped with LiDAR Sensor, UWB Tag, and Reflectors for the Motion Capture System. (b) \textit{Aisle} Environment at the Institute for Technical Logistics. (c) Recorded Contour Map of the \textit{Aisle} Environment. The Experiment Path, the Position of the LiDAR Reflectors, and the UWB Anchors are marked.}
\label{fig:setup}
\end{figure*}

\subsubsection{LiDAR ILS and Influencing Factors}
The LiDAR ILS computes the sensor's position via map matching. The distance of the sensor to objects in the environment is determined by the emission and detection of laser impulses. The position of the sensor is then determined by comparing the scan points with a previously recorded map of the environment applying a particle filter, supported by the data collected from an Inertial Measurement Unit (IMU). The influencing factors and parameter values are summarized in Table~\ref{tab:influence} on the left. Reflectors can be used as reference points to improve localization accuracy. Besides the presence of reflectors, the quality of the pre-recorded map, the dynamics of surrounding entities, and the Field of View (FoV) are considered application-driven influencing factors. In the experiments, the FoV is limited to either $270^{\circ}$ or $180^{\circ}$.  A person walking in front of the robot at a distance of 1.5 to 2.5m emulates dynamics. To account for map quality as an influencing factor, two environments are set up - the empty test area (\textit{Empty}) and an area with logistic objects representing a warehouse aisle (\textit{Aisle}). A map is recorded for each of the environment configurations. Figure~\ref{fig:setupb} shows the robot in the \textit{Aisle} environment, while Figure~\ref{fig:setupc} shows a recorded map in the same environment. In addition, the planned path, the positions of the LiDAR reflectors and UWB anchors are marked. The pairwise combination of the environments and maps results in four configurations. We quantify the map quality by recording an additional map during an exemplary experiment execution for each environment and comparing it with the original maps using a point cloud matching algorithm based on Iterative Closest Point (ICP) registration with a prior global registration step using the implementation in the python library Open3D \cite{Zhou2018}. Table~\ref{tab:map} presents the results of this process as a value between 0 and 1 for each of the four configurations, where 1 represents a perfect match. 

\begin{table}[t!]
    \centering
    \begin{tabular}{l|l|l|l}
         \multicolumn{2}{c|}{LiDAR}  & \multicolumn{2}{c}{UWB} \\
         \hline
         Factors &  Values & Factors  & Values\\
         \hline
         Dynamics & yes, no & Dynamics & yes, no\\
         Reflector & on, off & EKF & on, off \\
         FoV & [$180^{\circ}$, $270^{\circ}$] & Environment & empty, aisle\\
         Map quality & \makecell[t]{[0.54, 0.81, \\ 0.84, 0.99]} &  & 
    \end{tabular}
    \caption{Parameter Values for Considered Influencing Factors}
    \label{tab:influence}
\end{table}

\begin{table}[ht]
    \centering
    \begin{tabular}{l|l|l}
         & \multicolumn{2}{l}{Environment}\\
        \hline
         Map & \textit{Empty} & \textit{Aisle} \\
        \hline
         \textit{Empty} & 0.84 & 0.54 \\
         \textit{Aisle} & 0.81 & 0.99
    \end{tabular}
    \caption{Map Quality}
    \label{tab:map}
\end{table}

\subsubsection{UWB ILS and Influencing Factors}
The UWB ILS is based on the time-difference-of-arrival method. Localization systems based on radio-frequency signals are generally prone to None-Line-of-Sight (NLoS) errors. The \textit{Aisle} environment leads to realistic NLoS for the UWB ILS. Therefore, the environment is considered as an influencing factor. Dynamics are likewise considered for the LiDAR and the UWB ILS. Finally, the system can be configured to filter the position using an Extended Kalman Filter (EKF). 

\subsection{Experiment Execution}

Combining the parameter values from Table~\ref{tab:influence} results in 32 possible scenarios for the LiDAR and eight scenarios for the UWB ILS. To evaluate the repeatability of an experiment and reduce noise effects, three experiments are performed for each scenario, resulting in a total of 120 experiments. The localization systems are synchronized with the reference system via Precision Time Protocol (PTP)~\cite{ptpd} and aligned prior to the experiments. 
For each experiment, the system and the environment are configured according to the scenario. The localization and reference data are then continuously recorded as the robot successively traverses 33 evaluation poses at an average speed of 0.3 m/s. The resulting path describes the shape of two rectangles as shown in Figure~\ref{fig:setupc}, with the start and end point in the center.

\subsection{Accuracy Evaluation}

\begin{figure*}[t!]
\centering
\begin{subfigure}{.4\textwidth}
  \centering
  \includegraphics[height=6.5cm]{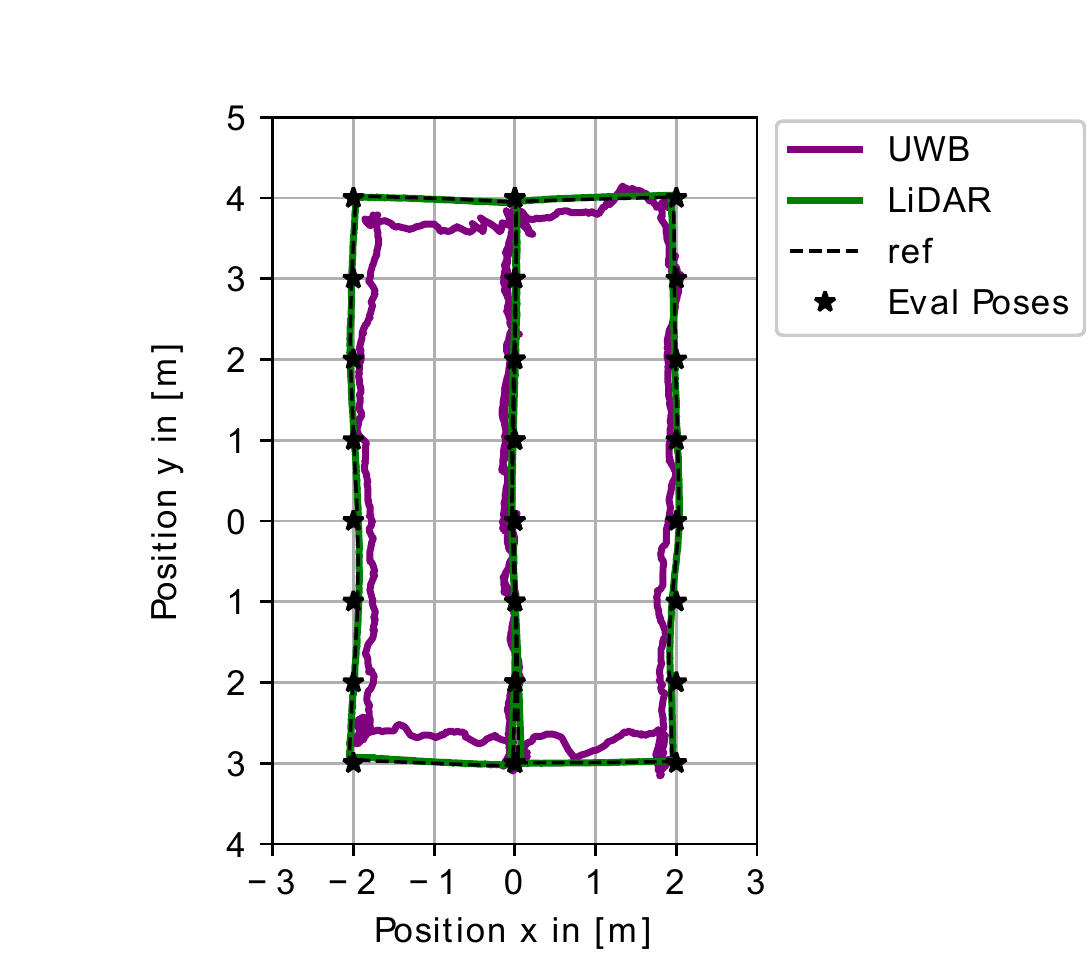}
  \subcaption{}
  \label{fig:evala}
\end{subfigure}
\begin{subfigure}{.59\textwidth}
  \centering
  \includegraphics[height=6.5cm]{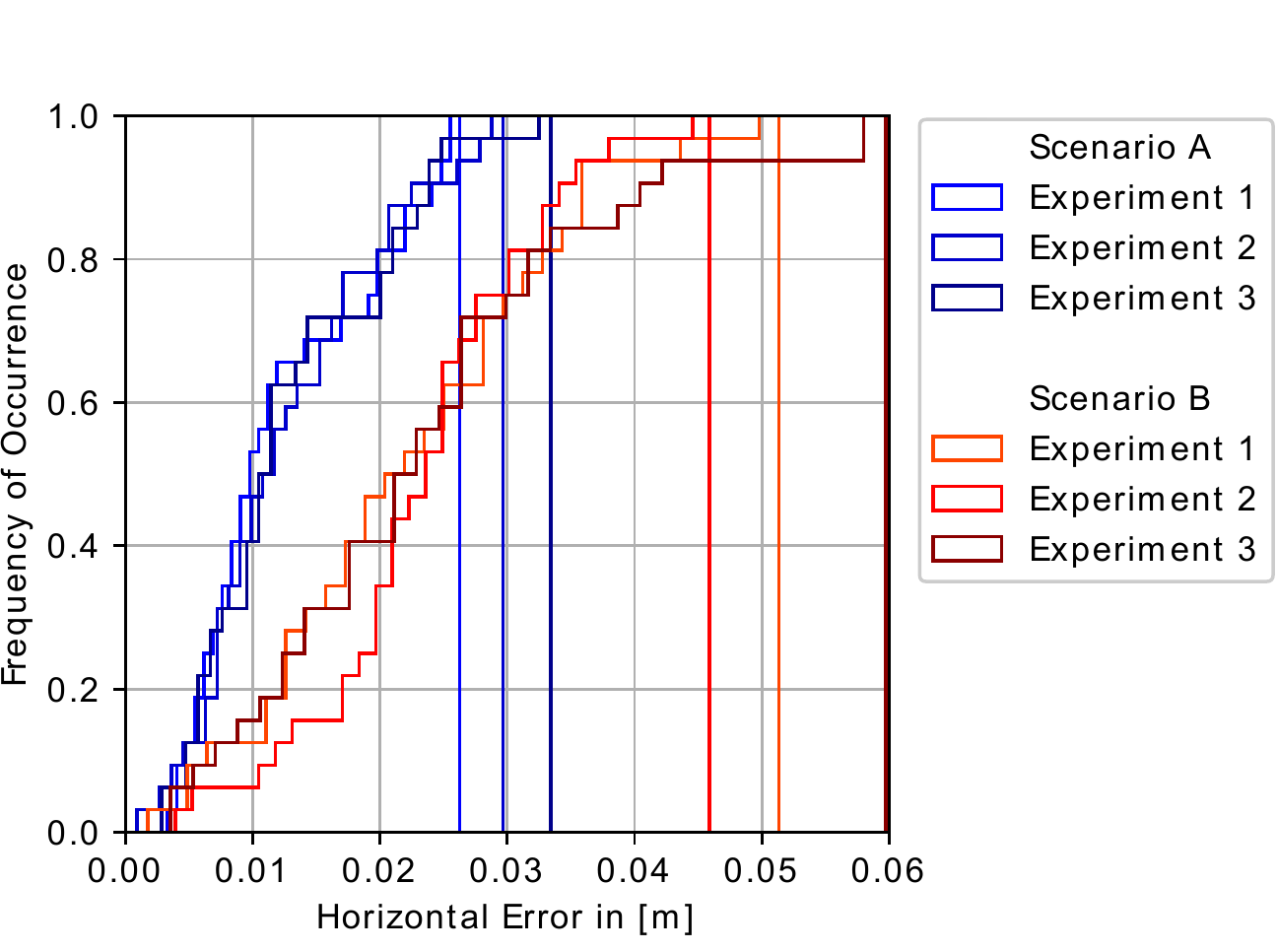}
  \subcaption{}
  \label{fig:evalb}
\end{subfigure}
\caption{Comparison of experiments: (a): UWB, LiDAR, and reference trajectories for a single experiment. (b): Cumulative distribution function for comparison of experiments and scenarios.}
\label{fig:eval}
\end{figure*}

The code provided with the T\&E~4iLoc Framework~\cite{s22072797} is used to compute the accuracy metrics for each experiment. Figure~\ref{fig:evala} shows the recorded trajectories of the UWB ILS, the LiDAR ILS, and the reference system for an exemplary experiment. Figure~\ref{fig:evalb} shows the cumulative distribution functions for six experiments from two scenarios for the LiDAR ILS. Comparing the experiments for one scenario indicates good repeatability, because their cumulative error frequencies are very similar, while the comparison of the scenarios indicates the significance of influences. The 95\%-percentile of the horizontal position error ($h_{95}$) at the evaluation poses is further considered as the performance metric. The median of the position error is an alternative, more robust metric, but high percentiles are more relevant for end-users and thus support an application-driven scenario. The mean value of the percentiles is used from the three experiments for each scenario.

\subsection{Output Categorization}

As described earlier, the categorization of the output can be either application-related or technology-related. Application-related categorization is favorable to determine the suitability of an ILS or its configuration for a specific application. Different processes of an application may require different localization accuracy. For the considered application of AMRs for warehouses, we define four performance classes for different processes with requirements for the 95\%-percentile of the horizontal position error. To enable the automated picking of goods, we require an accuracy of less then 0.05m (performance class A), while 0.1m is sufficient for material handling (performance class B). Tracking the absolute position of assets (performance class C) requires 0.5m and global navigation 1m (performance class D).

For technology-related categorization, automatic clustering via kNN is used together with the elbow criterion to determine the number of clusters.
For the data considered, the optimal number of clusters is five. Table~\ref{tab:out-cat} shows the clusters resulting from kNN for five clusters along with the results of application-related categorization. 

\begin{table}[!ht]
    \centering
    \small
    \begin{tabular}{c|c|c}
        \\
        Performance Class  & Requirements in [m] & Process\\
        \hline
        \multicolumn{3}{c}{
        \begin{minipage}{5cm}
        \centering
        Application-Related Categorization\\ (Based on Process Requirements)
        \end{minipage}
        } \\
        \hline
        A  & $ h_{95} <$ 0.05 & Goods Picking \\
        B  & 0.05 $\leq h_{95} <$ 0.1 & Material Handling \\
        C  & 0.1 $\leq h_{95} <$ 0.5 & Asset Tracking\\
        D  & 0.5 $\leq h_{95} <$ 1 & Navigation \\
        \hline
        \multicolumn{3}{c}{
        \begin{minipage}{5cm}
        \centering
        Technology-Related Categorization \\ (Based on System Capabilities)
        \end{minipage}
        } \\
        \hline
        I & $ h_{95}<$ 0.056 & -\\
        II  & 0.056 $\leq h_{95} <$ 0.209 & - \\
        III  & 0.209 $\leq h_{95} <$ 0.394 & - \\
        IV  & 0.493 $\leq h_{95} <$ 0.493 & - \\
        V  & $ h_{95} \geq  $ 0.493 & -
    \end{tabular}
    \caption{Performance Classes and Accuracy Requirements}
    \label{tab:out-cat}
\end{table}

Figure~\ref{fig:cluster} shows the 40 data points, i.e., the 95\%-precentiles of the horizontal position error for each scenario with the application-related and the technology-related performance classes. Intuitively, the automatic clustering provides reasonable categories. With the exception of one data point, the data points of performance classes A and B correspond to classes I and II, and the data points for performance class D correspond to class V. For technology-related categorization, the data points of performance class C are split into class III and IV.

\begin{figure}[ht!]
    \centering
    \includegraphics[width=\linewidth]{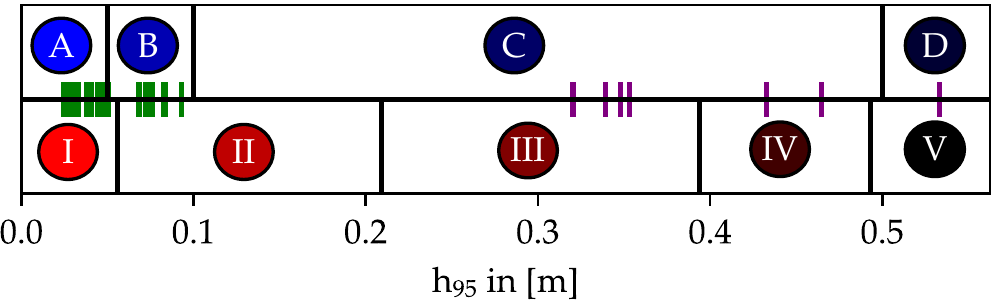}
    \caption{Data Points, combined for UWB (purple) and LiDAR ILS (green) with Application and Technology-Related Categorization}
    \label{fig:cluster}
\end{figure}

\subsection{Decision Trees}

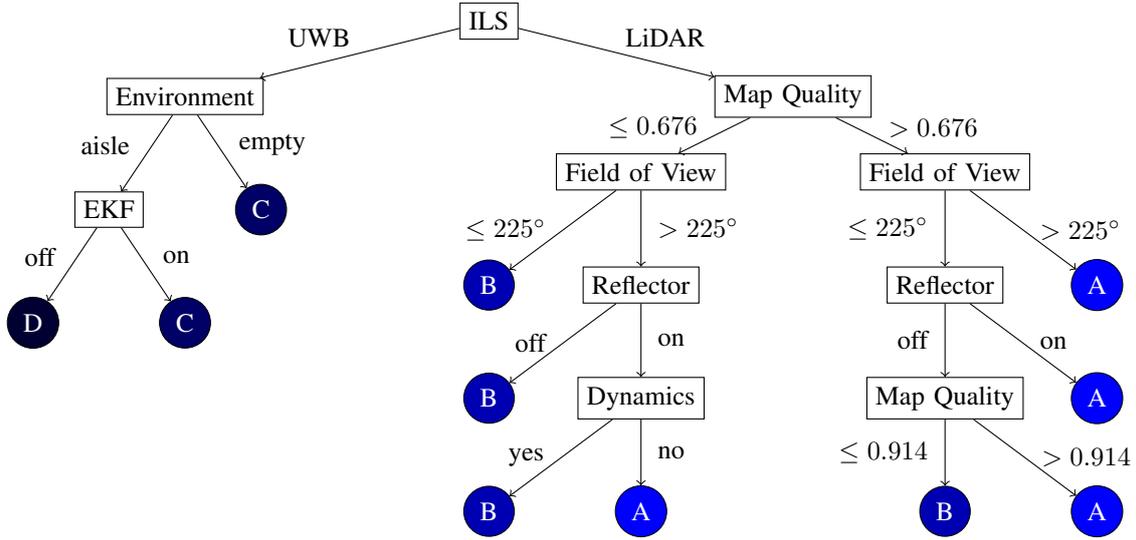
\begin{figure*}[t!]
    \centering
    \begin{tikzpicture}

\node[draw] at (0,0) (1) {ILS};

\node[draw,align=center] at (-4,-1) (2) {Environment};
\node[draw,align=center] at (4,-1) (3) {Map Quality};

\draw[->] (1) to node[left, yshift = 0.2cm] {UWB} (2);
\draw[->] (1) to node[right, yshift = 0.2cm] {LiDAR} (3);

\node[draw,align=center] at (-5,-2.5) (21) {EKF};
\node[draw,circle ,text=white,fill=black!60!blue, align=center] at (-3,-2.5) (26) {C};
\node[draw,align=center] at (2,-2) (4) {Field of View};
\node[draw,align=center] at (6,-2) (5) {Field of View};

\draw[->] (2) to node[right, xshift=0.1cm, yshift=0.1cm] {empty} (26);
\draw[->] (2) to node[left, xshift=-0.1cm, yshift=0.1cm] {aisle} (21);
\draw[->] (3) to node[left, xshift=-0.1cm, yshift=0.1cm] {$\leq 0.676$} (4);
\draw[->] (3) to node[right, xshift=0.1cm, yshift=0.1cm] {$> 0.676$} (5);

\node[draw,circle ,text=white,fill=black!80!blue, align=center] at (-6,-4) (23) {D};
\node[draw,circle ,text=white,fill=black!60!blue, align=center] at (-4,-4) (22) {C};
\node[draw,circle ,text=white,fill=black!30!blue, align=center] at (0,-3.5) (6) {B};
\node[draw,align=center] at (2,-3.5) (7) {Reflector};
\node[draw,align=center] at (6,-3.5) (8) {Reflector};
\node[draw,circle ,text=white,fill=black!0!blue, align=center] at (8,-3.5) (9) {A};

\draw[->] (21) to node[right, xshift=0.1cm, yshift=0.1cm] {on} (22);
\draw[->] (21) to node[left, xshift=-0.1cm, yshift=0.1cm] {off} (23);
\draw[->] (4) to node[left, xshift=-0.1cm] {$\leq 225^{\circ}$} (6);
\draw[->] (4) to node[right, xshift=0.1cm] {$> 225^{\circ}$} (7);
\draw[->] (5) to node[left, xshift=-0.1cm] {$\leq 225^{\circ}$} (8);
\draw[->] (5) to node[right, xshift=0.1cm] {$> 225^{\circ}$} (9);

\node[draw, circle ,text=white,fill=black!30!blue, align=center] at (0,-5) (10) {B};
\node[draw,align=center] at (2,-5) (11) {Dynamics};
\node[draw,align=center] at (6,-5) (12) {Map Quality};
\node[draw, circle ,text=white,fill=black!0!blue, align=center] at (8,-5) (13) {A};

\draw[->] (7) to node[left, xshift=-0.1cm] {off} (10);
\draw[->] (7) to node[right, xshift=0.1cm] {on} (11);
\draw[->] (8) to node[left, xshift=-0.1cm] {off} (12);
\draw[->] (8) to node[right,xshift=0.1cm] {on} (13);

\node[draw, circle ,text=white,fill=black!30!blue, align=center] at (0,-6.5) (14) {B};
\node[draw, circle ,text=white,fill=black!0!blue, align=center] at (2,-6.5) (15) {A};
\node[draw,  circle ,text=white,fill=black!0!blue, align=center] at (8,-6.5) (16) {A};
\node[draw, circle ,text=white,fill=black!30!blue, align=center] at (6,-6.5) (17) {B};

\draw[->] (11) to node[left, xshift=-0.1cm] {yes} (14);
\draw[->] (11) to node[right,xshift=0.1cm] {no} (15);
\draw[->] (12) to node[right, xshift=0.1cm] {$> 0.914$} (16);
\draw[->] (12) to node[left, xshift=-0.1cm] {$\leq 0.914$} (17);

\end{tikzpicture}
    \caption{Decision Tree for Application-Related Performance Classes}
    \label{fig:dt_app}
\end{figure*}

\begin{figure*}[th]
    \centering
    \begin{tikzpicture}

\node[draw] at (0,0) (1) {ILS};

\node[draw,align=center] at (-4,-1) (2) {Environment};
\node[draw,align=center] at (4,-1) (3) {Map Quality};

\draw[->] (1) to node[left, yshift = 0.2cm] {UWB} (2);
\draw[->] (1) to node[right, yshift = 0.2cm] {LiDAR} (3);

\node[draw,align=center] at (-6,-2) (21) {EKF};
\node[draw,align=center] at (-4,-2) (26) {EKF};
\node[draw,align=center] at (2,-2) (4) {Field of View};
\node[draw,align=center] at (6,-2) (5) {Field of View};

\draw[->] (2) to node[right, xshift=0.1cm, yshift=0.cm] {empty} (26);
\draw[->] (2) to node[left, xshift=-0.1cm, yshift=0.1cm] {aisle} (21);
\draw[->] (3) to node[left, xshift=-0.1cm, yshift=0.1cm] {$\leq 0.676$} (4);
\draw[->] (3) to node[right, xshift=0.1cm, yshift=0.1cm] {$> 0.676$} (5);

\node[draw,circle ,text=white,fill=black!100!red, align=center] at (-8,-3.5) (23) {V};
\node[draw,align=center] at (-6,-3.5) (22) {Dynamics};
\node[draw,circle ,text=white,fill=black!50!red, align=center] at (-2,-3.5) (40) {III};
\node[draw,align=center] at (-4,-3.5) (41) {Dynamics};
\node[draw,circle ,text=white,fill=black!25!red, align=center] at (0,-3.5) (6) {II};
\node[draw,align=center] at (2,-3.5) (7) {Reflector};
\node[draw,align=center] at (6,-3.5) (8) {Reflector};
\node[draw,circle ,text=white,fill=black!0!red, align=center] at (8,-3.5) (9) {I};

\draw[->] (21) to node[right, xshift=0.1cm, yshift=0.1cm] {on} (22);
\draw[->] (21) to node[left, xshift=-0.1cm, yshift=0.1cm] {off} (23);
\draw[->] (26) to node[right, xshift=0.1cm, yshift=0.1cm] {on} (40);
\draw[->] (26) to node[left, xshift=-0.1cm, yshift=0.1cm] {off} (41);
\draw[->] (4) to node[left, xshift=-0.1cm] {$\leq 225^{\circ}$} (6);
\draw[->] (4) to node[right, xshift=0.1cm] {$> 225^{\circ}$} (7);
\draw[->] (5) to node[left, xshift=-0.1cm] {$\leq 225^{\circ}$} (8);
\draw[->] (5) to node[right, xshift=0.1cm] {$> 225^{\circ}$} (9);

\node[draw, circle ,text=white,fill=black!50!red, align=center] at (-8,-5) (42) {III};
\node[draw, circle ,text=white,fill=black!75!red, align=center] at (-6,-5) (43) {IV};
\node[draw, circle ,text=white,fill=black!50!red, align=center] at (-4,-5) (44) {III};
\node[draw, circle ,text=white,fill=black!75!red, align=center] at (-2,-5) (45) {IV};
\node[draw, circle ,text=white,fill=black!25!red, align=center] at (0,-5) (10) {II};
\node[draw, circle ,text=white,fill=black!0!red, align=center] at (2,-5) (11) {I};
\node[draw,align=center] at (6,-5) (12) {Map Quality};
\node[draw, circle ,text=white,fill=black!0!red, align=center] at (8,-5) (13) {I};

\draw[->] (22) to node[left, xshift=-0.1cm] {off} (42);
\draw[->] (22) to node[right, xshift=0.1cm] {on} (43);
\draw[->] (41) to node[left, xshift=-0.1cm] {off} (44);
\draw[->] (41) to node[right, xshift=0.1cm] {on} (45);
\draw[->] (7) to node[left, xshift=-0.1cm] {off} (10);
\draw[->] (7) to node[right, xshift=0.1cm] {on} (11);
\draw[->] (8) to node[left, xshift=-0.1cm] {off} (12);
\draw[->] (8) to node[right,xshift=0.1cm] {on} (13);

\node[draw,  circle ,text=white,fill=black!0!red, align=center] at (8,-6.5) (16) {I};
\node[draw, circle ,text=white,fill=black!25!red, align=center] at (6,-6.5) (17) {II};

\draw[->] (12) to node[right, xshift=0.1cm] {$> 0.914$} (16);
\draw[->] (12) to node[left, xshift=-0.1cm] {$\leq 0.914$} (17);

\end{tikzpicture}
    \caption{Decision Tree for Technology-Related Performance Classes}
    \label{fig:dt_tech}
\end{figure*}
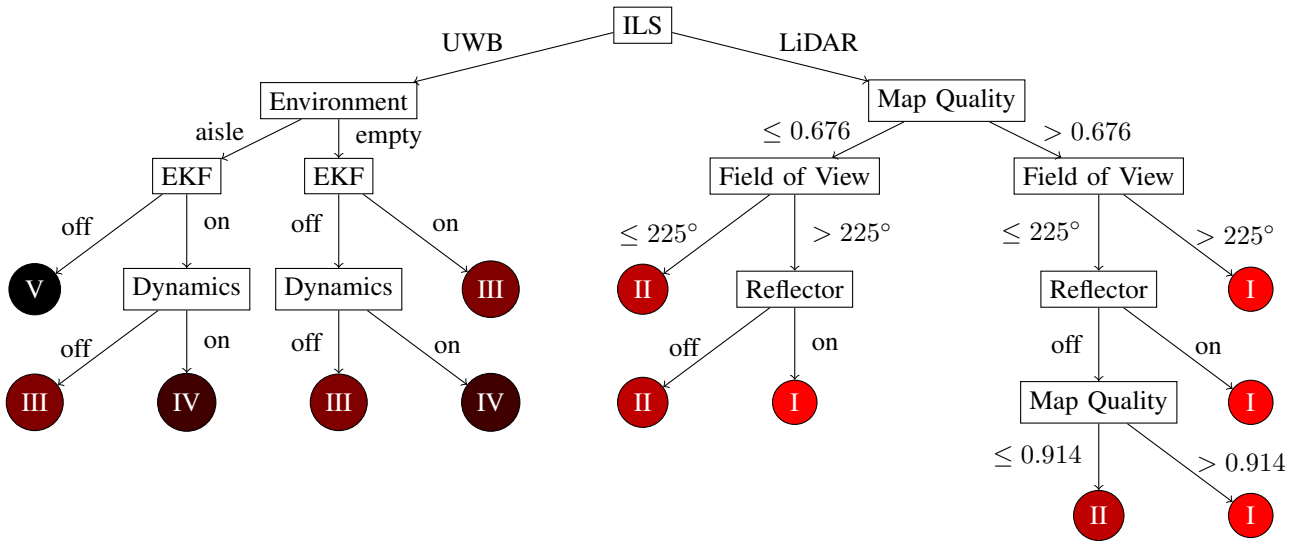

The python package scikit-learn \cite{scikit-learn} is used to learn the decision trees for the combined data set of the UWB and LiDAR ILS. We present the resulting decision trees for application-related and technology-related categorization in Figure~\ref{fig:dt_app} and Figure~\ref{fig:dt_tech}.  
Each leaf contains at least one data point. Since we use the average of three experiments, we have only one data point per feature vector. Therefore, there are no contradicting outcomes. 
For continuous data such as map quality or the FoV, the midpoint of the parameter values is chosen as the decision boundary, i.e., 0.676 and 225$^\circ$, respectively. The scenario is specified more and more following the tree-decisions from top to bottom. If an influence is not considered following a path from the root to a leaf of the tree, this influence is not relevant for the performance class represented by the leaf, i.e., all specifications for this influence result in an identical performance class.

We first consider the application-related decision tree from Figure~\ref{fig:dt_app}. 
The root node decides the type of system. For the UWB ILS, either performance class C or D are achievable, whereby the type of environment and the EKF support are relevant influences. Hence, the dynamics are not considered relevant, when deciding on a performance class. To achieve performance class C and thus meet the defined requirements for asset tracking, there are two options. Either the environment is \textit{Empty} (i.e., optimal LoS), or for the \textit{Aisle} environment the EKF support must be switched on. Following the branch of the LiDAR ILS, either performance class A or performance class B is reached, whereby all parameters are basically relevant. If the map quality is greater than 0.676, performance class B is only reached if the FoV is below 225$^\circ$, reflector support is switched off, and the map quality remains below 0.914. On the other hand, if the map quality is less than 0.676, performance class A is reached only if the FoV is greater than 225$^\circ$, reflector support is on, and there are no dynamics induced. 

The structure of the technology-related decision tree from Figure~\ref{fig:dt_tech} is similar to the application-related decision tree, which is an intuitive result considering the intersections of performance classes from Figure~\ref{fig:cluster}. On the LiDAR side, the only difference is that the dynamics are not relevant for any decision. This can be explained by the division of data points into categories presented earlier. There is one data point that belongs to performance class I and B, which results from the scenario \textit{'ILS: LiDAR,  Map Quality: 0.54, FoV: 270$^\circ$, Reflector: on, Dynamics: yes'}. When dynamics is on, the accuracy is reduced. However, for the technology-related categorization, this does not result in a worse performance class. On the UWB branch side, a new performance class is introduced, which leads to additional relevant features for determining scenarios from that performance class. If the \textit{Aisle} environment is chosen and EKF support is enabled, the induced dynamics is a decisive criterion. When the dynamics is turned off, the system performs better and performance class III is reached.
\section{Opportunities and Limitations of Analyzing Decision Trees}
\label{sec:discussion}

We introduced a new analysis based on the combination of a clearly determined T\&E process combined with meaningful output categorization (application- or technology-related) and the visualization using decision trees. In the following, we present opportunities and limitations of analyzing decision trees that represent influences on the accuracy of ILS according to our proposed strategy.

\subsection{Opportunities}

The decision tree representation automatically prepares and provides the experimental results in a simple structure that is intuitively interpreted by a human.

\paragraph{Increasing reproducibility and comparability for benchmarking ILS}
Ideally, the test environment can be abstracted by various influencing factors. When benchmarking ILS, the influencing factors on the system performance are often not clearly stated. By reporting the performance metrics not just as a single number, but by presenting the results with influences in the form of decision trees, the reproducibility and comparability of benchmarking results is increased. 

\paragraph{Understanding the impact of influencing factors on system performance}
Both users and system providers can use a decision tree to decide which influencing factors are relevant under which conditions. Looking at the application-related decision tree from Figure~\ref{fig:dt_app}, if the map quality for the LiDAR ILS is greater than 0.676 and the FoV is greater than 225$^\circ$, performance class A is achieved regardless of the reflector support.

\paragraph{Finding a suitable system for a given application}
If the requirements for an application including its environment are known, decision trees can be used to find the optimal system. In the given example, the process of picking goods would require the LiDAR ILS, while the UWB ILS would be sufficient for navigation.

\paragraph{Optimizing the system or environment configuration to meet application requirements}
In the event that the application requirements are not met by an ILS under given influences, the system and environment configuration can be optimized to reach a scenario that results in a satisfactory system performance. For example, for the scenario  \textit{'ILS: LiDAR,  Map Quality: 0.54, FoV: 270$^\circ$, Reflector: off, Dynamics: no'}, the use of reflectors would result in a change from performance class B to performance class A, enabling the process of good picking.

\subsection{Limitations}

Influences can be challenging to control, quantify/categorize, or generalize. For example, the influence of the recorded map on the accuracy of a LiDAR ILS can hardly be generalized by a single metric such as the suggested map quality value. Furthermore, continuous feature values in particular should be viewed with caution. For example, a FoV of 5$^\circ$ or a map quality of 0.1 could probably not end up in performance class A, as suggested by the decision trees in Figure~\ref{fig:dt_app} and Figure~\ref{fig:dt_tech}. In the presented case-study, no data is provided with FoV less than 180$^\circ$ or a map quality of less than 0.54. Other approaches for analyzing influencing factors, such as a factor analysis \cite{reyment_jvreskog_1993}, eventually, yield more accurate results for these cases. However, by using decision tree learning, we can model even non-linear dependencies and categorical influences. In addition, only few influencing factors were examined so far.  Further influences could be investigated, but it should be noted that the condition of testing the scenarios in all constellations leads to an exponential increase in the number of experiments.

In this case study, we collected data in a certain test hall under various restrictions, such as the used robot or path. For the potential of the analysis of influences with decision trees to unfold, it ultimately comes down to the quality of the collected data. To reliably enable the presented opportunities in practice, the T\&E data must represent the real world. Ideally, the experiments and influences reflect reality. In addition, the experiments should produce the same results when repeated multiple times in the same or a different test environment. Designing system-level experiments and defining influences that provide repeatable results that are relevant in real-world applications is a major challenge. 
\section{Conclusions}
\label{sec:conclusion}

In this work, a strategy for determining the position accuracy of ILS under application-related influences and the representation of the data in decision trees was presented. In a first step, the absolute position accuracy has to be determined by empirical test and evaluation. The T\&E 4iLoc framework is used as methodology for test and evaluation.. The framework provides a systematic way to determine performance metrics for application-dependent test scenarios consisting of various influencing factors. Next, the performance metrics from several experiments under different influences are categorized into performance classes. This can either be done application-related based on application requirements, or technology-related, based on automatic clustering. For automatic clustering, we propose the kNN algorithm in combination with an elbow criterion. Finally, the experiments associated with the influencing factors are labelled with the performance classes and decision trees are learned. 

The strategy was successfully demonstrated by empirical experiments. The horizontal position accuracy ($h_{95}$) of a UWB and a LiDAR ILS was determined considering the  application of AMRs in warehouses for a total of 120 experiments under different influences. The application is considered to define application-driven influencing factors and application-related output categorization. The decision trees provide a simple and intuitively interpretable overview of the systems's performance, depending on the influences. Furthermore, the opportunities of analyzing decision trees for benchmarking ILS, identifying a suitable system, optimizing the configuration, and understanding the impact of influencing factors are presented. The opportunities of analyzing the decision trees in practice ultimately depend on the quality of the T\&E results. Nonetheless, our proposed strategy provides a simple way to generate decision trees as an intuitive, comprehensible and meaningful representation of the influences on the position accuracy of ILS. 

\section*{Acknowledgment}
This work is supported by the TUHH $I^3$ Projects funding. Special thanks go to the 3D\_Log project team. 

\begin{spacing}{1.05}
    \footnotesize
    \bibliographystyle{IEEEtranN}
    \bibliography{bibliography.bib}
\end{spacing}

\end{document}